\documentclass[graybox]{svmult}

\usepackage{type1cm}        
                            
\usepackage{makeidx}         
\usepackage{graphicx}        
                            
\usepackage{multicol}        
\usepackage[bottom]{footmisc}

\usepackage{hyperref} 

\usepackage{newtxtext}       
\usepackage[varvw]{newtxmath}       

\usepackage{subcaption}

\makeindex            
\begin{document}

\title*{Pseudo-Boolean Polynomials Approach To Edge Detection And Image Segmentation}

\author{Tendai Mapungwana Chikake, Boris Goldengorin and Alexey Samosyuk}

\institute{
Tendai Mapungwana Chikake 
\at 
    Department of Discrete Mathematics, 
	Phystech School of Applied Mathematics and Informatics, 
	Moscow Institute of Physics and Technology, 
	Institutsky lane 9, 
	Dolgoprudny, 
	Moscow region, 141700, Russian Federation
	\email{tendaichikake@phystech.edu}
\and 
Boris Goldengorin \at
	Department of Mathematics, 
	New Uzbekistan University, 
	Tashkent, 
	100007, 
	Uzbekistan.
	The Scientific and Educational Mathematical Center “Sofia Kovalevskaya Northwestern Center for Mathematical Research” in Pskov State University, 
	Sovetskaya Ulitsa, 21, Pskov, Pskovskaya oblast’, 180000, Russian Federation.
	Department of Discrete Mathematics, 
	Moscow Institute of Physics and Technology, 
	Russian Federation, 
	\email{goldengorin.bi@mipt.ru}
\and
Alexey Samosyuk 
\at 
    Department of Discrete Mathematics, 
    Phystech School of Applied Mathematics and Informatics, 
    Moscow Institute of Physics and Technology, 
    Institutsky lane 9, 
    Dolgoprudny, 
    Moscow region, 141700, Russian Federation,
    \email{alexeysamosyuk@phystech.edu}
}

\maketitle

\abstract*{
    We introduce a deterministic approach to edge detection and image segmentation by formulating pseudo-Boolean polynomials on image patches.
    The approach works by applying a binary classification of blob and edge regions in an image based on the degrees of pseudo-Boolean polynomials calculated on patches extracted from the provided image. 
    We test our method on simple images containing primitive shapes of constant and contrasting colour and establish the feasibility before applying it to complex instances like aerial landscape images. 
    The proposed method is based on the exploitation of the \textit{reduction}, polynomial degree, and \textit{equivalence} properties of penalty-based pseudo-Boolean polynomials.
}

\abstract{
    We introduce a deterministic approach to edge detection and image segmentation by formulating pseudo-Boolean polynomials on image patches.
    The approach works by applying a binary classification of blob and edge regions in an image based on the degrees of pseudo-Boolean polynomials calculated on patches extracted from the provided image. 
    We test our method on simple images containing primitive shapes of constant and contrasting colour and establish the feasibility before applying it to complex instances like aerial landscape images. 
    The proposed method is based on the exploitation of the \textit{reduction}, polynomial degree, and \textit{equivalence} properties of penalty-based pseudo-Boolean polynomials.
}

\section{Introduction}

In digital image processing and computer vision, image segmentation is the process of partitioning a digital image into multiple image segments, also known as image regions or image objects \cite{shapiro_computer_2001}.
The process has close relationship to blob extraction, which is a specific application of image processing techniques, whose purpose is to isolate (one or more) objects (aka. regions) in an input image \cite{yusuf_blob_2018}.

The goal of segmentation is to simplify and/or change the representation of an image into something that is more meaningful and easier to analyse \cite{shapiro_computer_2001}.

There exist classical and AI based methods for the purpose of segmentation/blob-extraction. 
These methods converge into semantic\cite{guo_degraded_2020}, instance\cite{yi_attentive_2019} and panoptic\cite{kirillov_panoptic_2019} segmentation.
The underlying techniques of these methods can be classified into threshold filtering, clustering, differential motion subtraction, histogram, partial-differential equation solving, graph partitioning, supervised neural-network association and edge detecting methods. 
Our proposed method lies at the intersection of edge detection and threshold filtering methods.

Image segmentation plays a pivotal role, as a preprocessing step in localizing region of interest in content-based image retrieval, anomaly detection in industrial imagery, medical imaging, object detection and recognition tasks.

Our proposed method utilise the \textit{reduction}, \textit{equivalence} and \textit{degree} properties of penalty-based pseudo-Boolean polynomials for purposes of extracting regions of interest which provide course segmentation that can be extended to image segmentation. 
We derive our methodology from \cite{albdaiwi_data_2011}'s work on penalty-based pseudo-boolean formulation for purposes of data aggregation on p-median problems in the context of image processing.  

Our method bases its operation on a combination of threshold filtering and edge detection by deterministically grouping masks which convey regions of colour gradient shift.

Localizing regions of interest in a blind manner is a difficult task, often requiring pattern recognition of big image data to derive useful utility.
Our approach avoids learning of patterns from data and operate in a deterministic manner.

In our method, we partition an input image into small patches which we treat as \textit{information cost} matrices. 
We aggregate these matrices into their smallest possible pseudo-Boolean polynomials and group together equivalent instances, thereby delineating spatially contrasting regions in image representations. 

We verify our method by applying it to simple images containing primitive shapes and then scale up to simple natural scene images where we show that the method can competitively extract segments.

Our proposed method seeks to
\begin{itemize}
        \item introduce a blind preprocessing step for semantic segmentation
        \item assist with unsupervised annotation of segmentation datasets
        \item promote deterministic approaches to computer vision solutions.
\end{itemize}

The proposed method requires tuning of three parameters to balance performance and fine-tuned segmentation results.

\section{Related work}

Currently, our proposed method acts as a supportive step to a final segmentation processor.
This means that our method is still a complementary processing tool to complete an image segmentation task. 

Complete semantic segmentation requires the clustering of parts of an image together and proposing an object class while instance segmentation is concerned with detecting and delineating each distinct object of interest appearing in an image \cite{chennupati_learning_2021}. 

Our current solution cannot yet attach object classes but can distinctly delineate some individual objects in an image. 
The method achieves this by separating neighbouring pixels into blob or edge region based on the degree of a pseudo-Boolean polynomial calculated on patches extracted from the image. 

The resulting masks can be refined into classifications or delineations of distinct objects by an additional process which we are still working on. 
Our end goal is to achieve a solution for complete segmentation.

By the time of writing, popular solutions to instance and semantic segmentations are mostly based on either Mask R-CNN \cite{heMaskRCNN2018d} or the U-Net Convolutional Network \cite{ronneberger_u-net_2015}. 
Both solutions are learning-based methods and require large amounts of labelled ground truth data. 
Neural network based methods also require accelerated computer processors to perform in reasonable time.
Our solution avoids learning rules from data and can be easily run on non-accelerated CPUs.

Since our method is based on classifying edge and blob regions in an image, the Canny edge detector \cite{Canny}, introduced in 1986 by John F. Canny, is a closely related technique.
The Canny edge detector is an edge detection operator which uses a multi-stage algorithm to detect a wide range of edges in images. 
The Canny method detects edges by first applying a Gaussian filter to smooth the image in order to remove the noise, followed by the finding of the intensity gradients in the image \cite{liu_image_2017}.
After these steps, the method applies a gradient magnitude threshold filtering or lower bound cut-off suppression to get rid of spurious response to edge detection and then apply a double threshold to determine potential edges and conclude its process by tracking edges by hysteresis \cite{Canny}.
The Canny algorithm is adaptable to various environments because its parameters allow it to be tailored to recognition of edges of differing characteristics depending on the particular requirements of a given implementation \cite{liu_image_2017}. 
The optimised Canny filter is recursive, and can be computed in a short, fixed amount of times, but the implementation of the Canny operator does not give a good approximation of rotational symmetry and therefore gives a bias towards horizontal and vertical edges \cite{liu_image_2017}. 
In contrast, our method, calculate the pseudo-Boolean polynomial in normal and transposed patch matrices and selects the pseudo-Boolean polynomial form with the highest degree during the blob/edge classification step, thereby avoiding the edge direction bias. 

From an indirect perspective we can look at blind aggregation of possible distinct objects in image data in terms of blob extraction. 
Popular solutions using this paradigm, have underlying usage of either of the popular blob extraction methods which include Laplacian of Gaussian \cite{Laplacian_of_Gaussian}, Difference of Gaussian \cite{assirati_performing_2014}, or Determinant of a Hessian \cite{li_pattern_2014}, among other methods.

Laplacian of Gaussian is a blob extraction method which determines the blobs by using the Laplacian of Gaussian filters \cite{Laplacian_of_Gaussian}. 
The Laplacian is a 2-D isotropic measure of the 2nd spatial derivative of an image which highlights regions of rapid intensity change and is therefore, often used for edge detection \cite{Laplacian_of_Gaussian}. 
We often apply the Laplacian after an image smoothening method with something approximating a Gaussian smoothing filter in order to reduce its sensitivity to noise. 
In our method, the Gaussian smoothing filter step is an optionally used when applied to noisy image instances. 

The Difference of Gaussian method determines blobs by using the difference of two differently sized Gaussian smoothed images and follows generally most of the concept of the Laplacian of Gaussian \cite{Laplacian_of_Gaussian}. 

Determinant of a Hessian to aggregate regions of possible segmentation is achieved by determining blobs using the maximum in the matrix of the Hessian determinant \cite{li_pattern_2014}.

Generally, blob extraction based methods, propose small and numerous regions of interest making them hard to extend into spatial segmentation.

On the other hand, our method aggregates equivalent regions, by assigning zero or pseudo-Boolean polynomials with lower degree as blob regions and edge otherwise.
Regions initially described in different colour distributions in the pixel array often output high-order pseudo-Boolean polynomials which indicate contour regions.
This property allows us to extract larger and fewer spatial regions of interest in an image, making our method stand out against the other blob aggregation methods.

\section{Methods}

In mathematics and optimization, a pseudo-Boolean function is a function of the form ${f: \mathbf{B^n} \to \mathbb{R}}$, where ${\mathbf{B} = \{0, 1\}}$ is a Boolean domain and ${n}$ is a non-negative integer called the degree of the function.  \cite{boros_pseudo_boolean_2002}

\cite{goldengorin_cell_2013} highlights fundamental reduction and equivalence properties of penalty-based pseudo-Boolean polynomials which guarantee the maintenance of underlying initial information while reducing the problem complexity. 

We utilise this formulation on image patches to achieve compact representations of patches, and based on the \textit{degrees} and \textit{equivalence}\cite{goldengorin_cell_2013} properties of the resultant pseudo-Boolean polynomials, we can qualify an edge/blob classifier on input patches. 

Given an image patch represented by an ${m \times n}$ sized matrix which we treat as an information cost matrix ${C}$, our first task is to determine the minimum possible way of representing this cost in a way that allows us to compare if the patch is extracted from a blob region or a contour region. 

A patch that overlaps regions of contrasting information (i.e. overlapping an edge) in an image, results in a representation that is costly in comparison to one that lies over a blob region.
We claim this assertion because matrix values in blob regions are usually equivalent or have very small differences between each other and since \cite{goldengorin_cell_2013}'s pseudo-Boolean polynomial formulation is penalty-based, similar costs cancel out.

Reducing the patches to their smallest pseudo-Boolean polynomials provides comparable instances which can be tested for equivalence as well.
This property is termed \textit{equivalence} in \cite{goldengorin_cell_2013} and can be used to fine-tune edges or detect similar regions on the image matrix. 

The Pseudo-Boolean representation according to \cite{albdaiwi_data_2011} requires the generation of a coefficients matrix and its respective terms' matrix, whose combination creates monomials of the pseudo-Boolean polynomial. 

After the generation of these matrices, most of the processing: local aggregation, reduction of columns, and truncation processes are heavily dependent on the terms' matrix. 

Using an example instance to illustrate the formulation process, we take a ${4 \times 5}$ patch from an image. 

Let 
\[
C = 
    \begin{bmatrix}
        8 & 8 & 8 & 5\\
        12 & 7 & 5 & 7\\
        18 & 2 & 3 & 1\\
        5 & 18 & 9 & 8\\
    \end{bmatrix}
\]

The terms encoding function takes as input the permutations (${\Pi}$) matrix which is an index ordering of the input matrix.
\[
	\Pi = 
    \begin{bmatrix}
        4 & 3 & 3 & 3\\
        1 & 2 & 2 & 1\\
        2 & 1 & 1 & 2\\
        3 & 4 & 4 & 4\\
    \end{bmatrix}
\]

Using ${\Pi}$, we sort the initial cost matrix ${C}$ 

\[
	\text{sorted}\, C = 
    \begin{bmatrix}
        5 & 2 & 3 & 1\\
        8 & 7 & 5 & 5\\
        12 & 8 & 8 & 7\\
        18 & 18 & 9 & 8\\
    \end{bmatrix}
\]

and derive the ${\Delta C}$ matrix
\[
	\Delta C  = 
    \begin{bmatrix}
        5 & 2 & 3 & 1\\
        3 & 5 & 2 & 4\\
        4 & 1 & 3 & 2\\
        6 & 10 & 1 & 1\\
    \end{bmatrix}
\]

Using the ${\Pi}$ matrix we calculate the terms' matrix

\[
	\mathbf{y} =  
	\begin{bmatrix}
        \\
        y_{4} & y_{3} & y_{3} & y_{3}\\
        y_{1}y_{4} & y_{2}y_{3} & y_{2}y_{3} & y_{1}y_{3}\\
        y_{1}y_{2}y_{4} & y_{1}y_{2}y_{3} & y_{1}y_{2}y_{3} & y_{1}y_{2}y_{3}\\
    \end{bmatrix}
\]

and derive the resulting pseudo-Boolean polynomial

\[
	\begin{bmatrix}
        5 & 2 & 3 & 1\\
        3y_{4} & 5y_{3} & 2y_{3} & 4y_{3}\\
        4y_{1}y_{4} & 1y_{2}y_{3} & 3y_{2}y_{3} & 2y_{1}y_{3}\\
        6y_{1}y_{2}y_{4} & 10y_{1}y_{2}y_{3} & 1y_{1}y_{2}y_{3} & 1y_{1}y_{2}y_{3}\\       
	\end{bmatrix}
\]

We then perform local aggregation by summing similar terms and get a compact representation of the initial instance as 

\[
    \begin{bmatrix}
        0 & 0 & 11\\
        0 & 11y_{3} & 3y_{4}\\
        2y_{1}y_{3} & 4y_{2}y_{3} & 4y_{1}y_{4}\\
        0 & 12y_{1}y_{2}y_{3} & 6y_{1}y_{2}y_{4}\\
    \end{bmatrix}
\]
which in this particular example has 50\% less cost compared to the initial instance when expressed as polynomial.

We then search for the equivalent matrix with the minimum number of columns and in this particular example, it is already reduced to this state.

When pseudo-Boolean polynomials are reduced to their smallest instance, an equivalence property is apparent because different instances of similar information converge into a similar reduced pseudo-Boolean polynomials. 

This property is central to the blob aggregation task, and in this task these regions of equivalence in the reduced pseudo-Boolean polynomials context, occupy the set of blobs.

Below are examples of cost matrices, which are initially different but converge into similar reduced instances.

\[
	\begin{bmatrix}
		138 & 138 & 138 & 136 \\
		139 & 139 & 138 & 137 \\
		142 & 141 & 139 & 138 \\
		142 & 140 & 139 & 138 \\
	\end{bmatrix}
\]

and 

\[
	\begin{bmatrix}
		136 & 136 & 138 & 140 \\
		138 & 137 & 138 & 140 \\
		140 & 139 & 140 & 141 \\
		139 & 139 & 140 & 141 \\
	\end{bmatrix}
\]

reduce to 

\[
	\begin{bmatrix}
		550\\
		3y_1\\
		6y_1y_2\\
		1y_1y_2y_4\\
	\end{bmatrix}
\]

\begin{figure}
    \caption{Pseudo-Boolean polynomial degree plotted for an image of primitive shapes}\label{fig:degree_illustration}
    \includegraphics[width=1.0\textwidth]{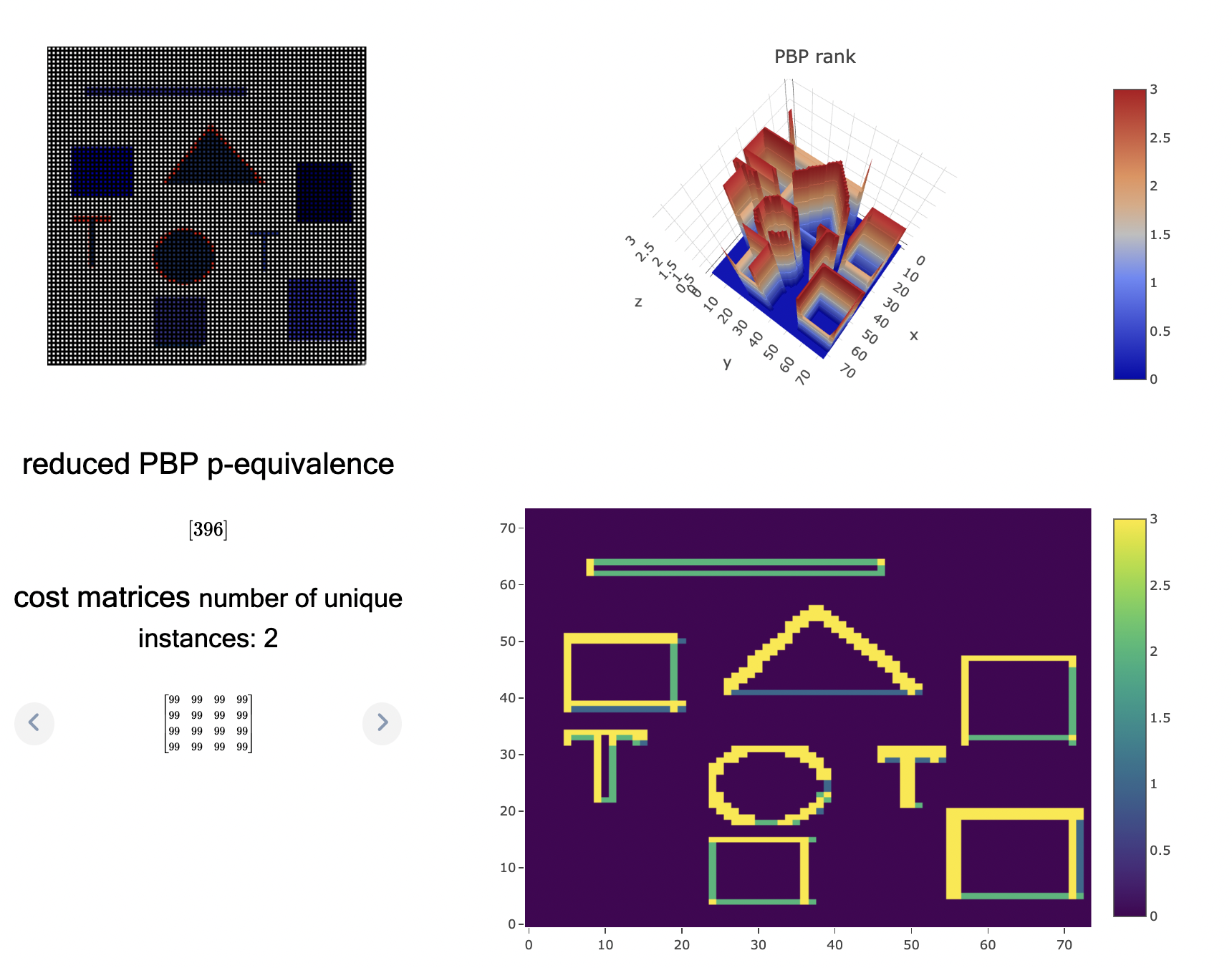}
    \centering
\end{figure}

By plotting the degree of the pseudo-Boolean polynomials of each patch on a surface plot, we observe contours available in spatial features of a given image as shown in Fig.~\ref{fig:degree_illustration}.

For our edge/blob classifier, we use the pseudo-Boolean polynomial degree ${r}$ to classify whether the particular patch lies over a contour or a blob region. 
We select ${p < m}$ to be the cut-off threshold for classification and based on this value, we can alter how fine/course should our edges be. 
If the degree ${r}$ of the pseudo-Boolean polynomial calculated on the patch is higher than ${p}$ then the patch lies over an edge or blob otherwise.
The edge/blob classifier if simply 

\begin{equation}
    f(r, p) =
    \begin{cases}
        \text{edge} & \text{if}\, r < p, \\
        \text{blob} & \text{otherwise}
    \end{cases}
\end{equation}

Given an image, broken into small patches of size ${4 \times 4}$, the maximum possible degree is ${3}$, and we can select ${p = 1}$, such that we have a binary set of patches, where patches, whose pseudo-Boolean polynomials reduce to a constant are described as blob regions, and the rest as edge points.
 
Using \textit{equivalence} \cite{albdaiwi_data_2011}, we can fine-tune the classified edges, should neighbouring patches exhibit contrasting edge/blob classes by equating the blobs in favour of the edge or vice-versa depending on how fine we require our edges to be. 

\section{Experimental Setup}

\begin{figure}
    \centering
    \begin{subfigure}{.4\textwidth}
        \centering
        \includegraphics[width=.8\linewidth]{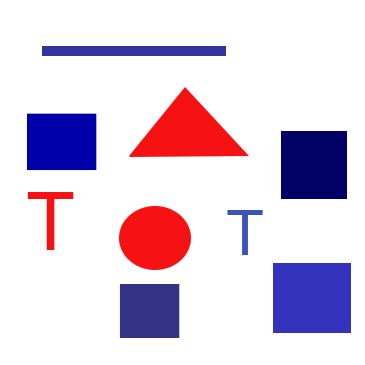}
        \caption{Input image}
        \label{fig:primitives_input}
    \end{subfigure}%
    \begin{subfigure}{.4\textwidth}
        \includegraphics[width=.8\textwidth]{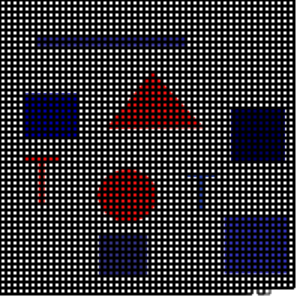}
        \centering
        \caption{Showing ${6 \times 6}$ sized patches}
        \label{fig:patched_primitives}
        \end{subfigure}
    \caption{Input image of primitives and its patching process}
    \label{fig:test}
\end{figure}

Given an image of size ${200 \times 200}$ containing basic primitive shapes of continuous colour shown in Fig.~\ref{fig:primitives_input}, we extract patches of significantly smaller sizes, e.g. ${6 \times 6}$ as shown in Fig.~\ref{fig:patched_primitives}.

We then apply the formulation and reduction of pseudo-Boolean polynomials on each patch and group them into a binary set ${S = \{\text{Blob}, \text{Edge}\}}$ based on the pseudo-Boolean polynomial degree.

Patches whose pseudo-Boolean polynomial degree ${r < p}$ are considered \textbf{Blobs} and \textbf{Edge} otherwise.

Patches with constant pixel values ${x \in [0, 255]}$ like 
\[
    \begin{bmatrix}
        99 & 99 & 99 & 99\\
        99 & 99 & 99 & 99\\
        99 & 99 & 99 & 99\\
        99 & 99 & 99 & 99\\
    \end{bmatrix}
\]
are guaranteed to converge to constant (zero-degree pseudo-Boolean polynomials), 

\[
    \begin{bmatrix}
        396\\
    \end{bmatrix}
\]

while some, with varied pixel values which cancel out like 
\[
    \begin{bmatrix}
        254 & 254 & 19 & 84\\
        254 & 254 & 19 & 84\\
        254 & 254 & 19 & 84\\
        254 & 254 & 19 & 84\\
    \end{bmatrix}
\]
also converge to constant (zero-degree pseudo-Boolean polynomials)
\[
    \begin{bmatrix}
        611\\
    \end{bmatrix}
\] 
Consequently these patches are grouped among the set of patches whose information cost is described as blob region. 

There are patches which contain varied pixel data which may converge to high-order pseudo-Boolean polynomials like 
\[
    \begin{bmatrix}
        254 & 254 & 6 & 17\\
        254 & 254 & 6 & 17\\
        254 & 254 & 6 & 17\\
        254 & 254 & 6 & 123\\
    \end{bmatrix}
\]
and 
\[
    \begin{bmatrix}
        254 & 254 & 6 & 17\\
        254 & 254 & 6 & 123\\
        254 & 254 & 6 & 123\\
        254 & 254 & 6 & 123\\
    \end{bmatrix}
\]

which converge to 
\[
    \begin{bmatrix}
        531\\
        106y_{1}y_{2}y_{3}\\
    \end{bmatrix}
\]
a 3rd-degree pseudo-Boolean polynomial which would be classified as lying on edge regions, should the cut-off threshold be given as ${r = 2}$.
In the particular simple case of the image in Fig.~\ref{fig:primitives_input}, patches which lie at shape edges converge to these non-zero degree pseudo-Boolean polynomials.

\begin{figure}
    \includegraphics[width=.4\textwidth]{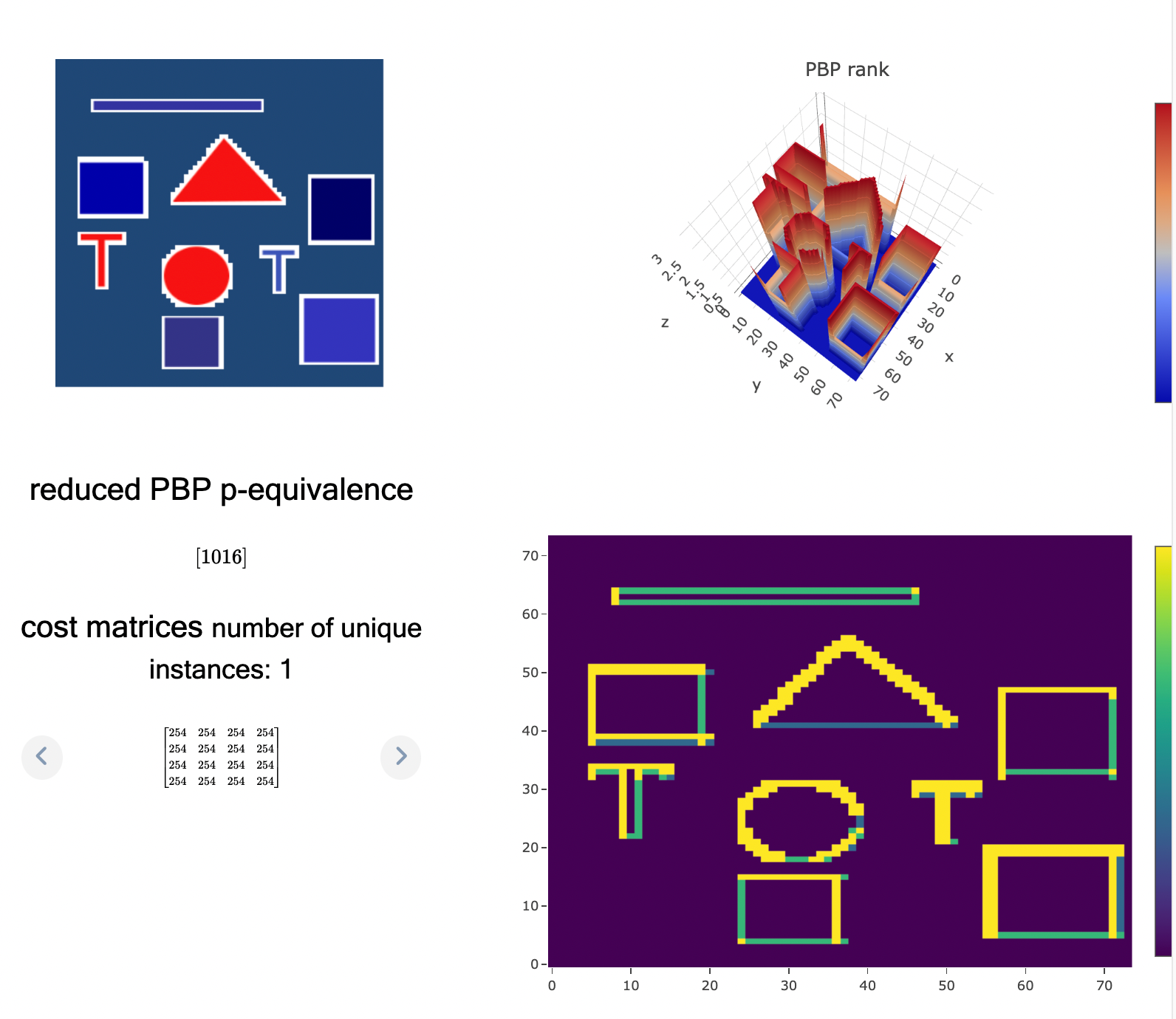}
    \centering
    \caption{colour coding \textit{blue} to discovered, zero-degreed pseudo-Boolean polynomials of input instances}\label{fig:zero-group}
\end{figure}

By colour-coding the patches which belong to the set of patches which have pseudo-Boolean polynomial's degree ${r = 0}$, we can observe regions of colour continuation, which we can classify as blobs as shown in Fig.~\ref{fig:zero-group}

As can be observed from Fig.~\ref{fig:zero-group}, the method allows us to find border points (white coloured) on simple objects found in the image. 

Applying the same process on a natural and unprocessed image may not result in desirable or useful aggregation as natural images often have less drastic transition of spatial features.
For this reason we ported the use of a Gaussian filter as a preprocessing step, followed by a pixel set aggregation step which is essentially a multivalued threshold processing.
Instead of raw pixels as input in our patches, we group ranges into \textit{pixel sets} whose size depends on the variance of pixel distribution in the image to promote group convergence of pseudo-Boolean polynomials. 

We create these ${ f(x) = \text{\textit{pixel sets}}}$ as 

\begin{equation}
    f(x) =
    \begin{cases}
        0 & \text{if}\, x < 5, \\
        1 & \text{if}\, x \in [5, 10), \\
        2 & \text{if}\, x \in [10, 15), \\
        \text{...} \\
        \text{...} \\
        \text{...} \\
        51 & \text{if}\, x > 250,
    \end{cases}
\end{equation}

thereby reducing the information cost range from ${[0, 255]}$ to ${[0, 51]}$ for instance.

Natural images tend to have smooth transitions of pixel values for neighbouring pixels at atomic level due to anti-aliasing in RGB representation of image data, thereby reducing the chances of neighbouring pixel patches converging into equivalent groups and consequently coarse edges but the Gaussian filter preprocessing together with the grouping of pixel ranges promotes finer edges.

\section{Results and Discussion}

\begin{figure}
    \includegraphics[width=.5\textwidth]{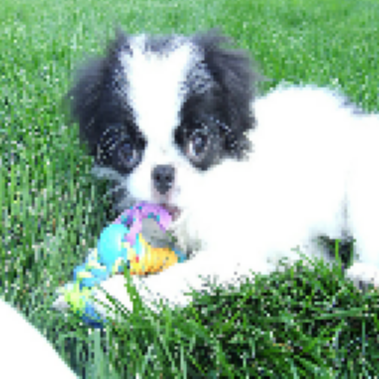}
    \centering
    \caption{Input natural image for preprocessing comparison }
    \label{fig:input_dog}
\end{figure}

\begin{figure}
    \includegraphics[width=.5\textwidth]{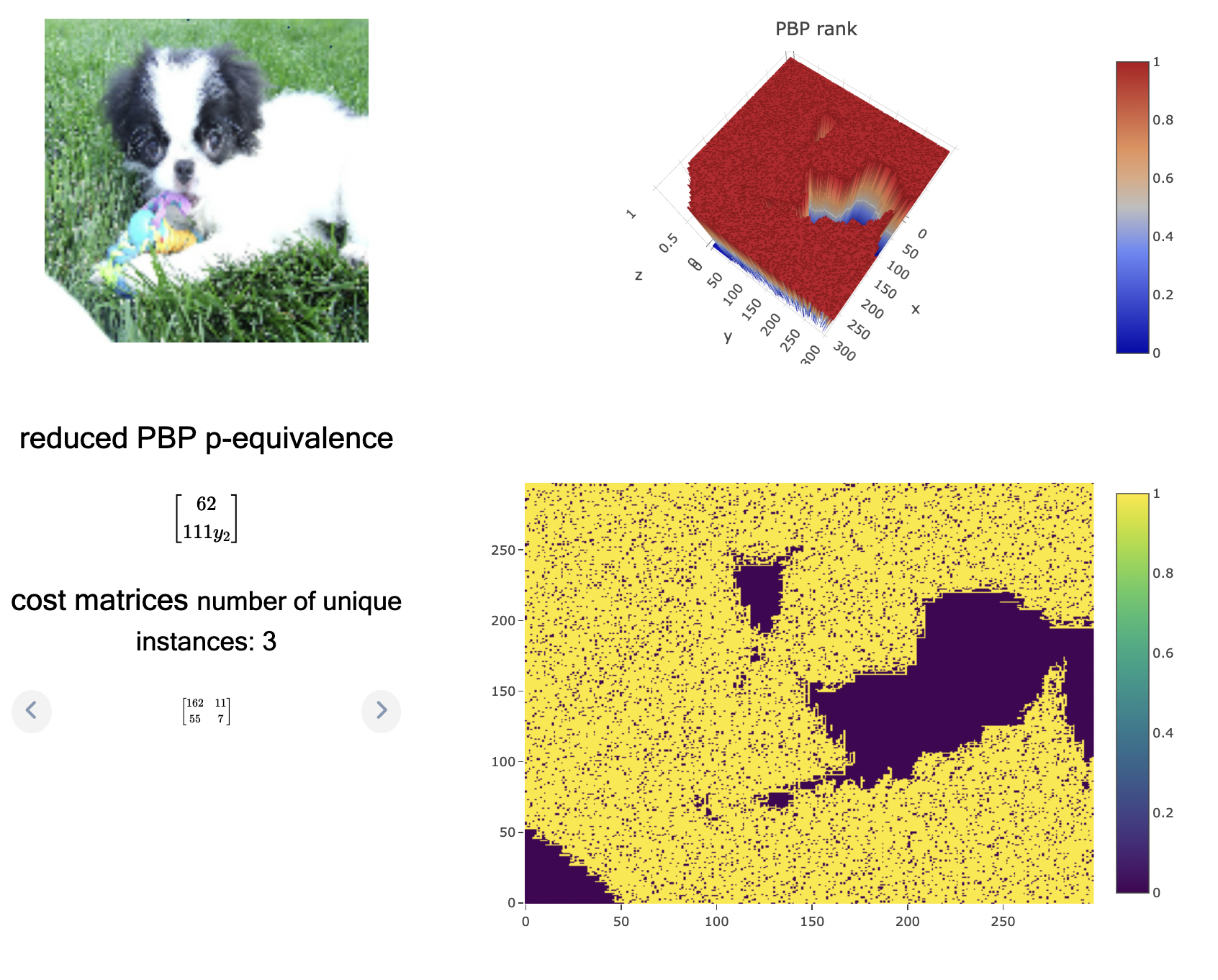}
    \centering
    \caption{Segmentation without the pre-processing steps }
    \label{fig:pbp_without_gaussian}
\end{figure}

\begin{figure}
    \includegraphics[width=.5\textwidth]{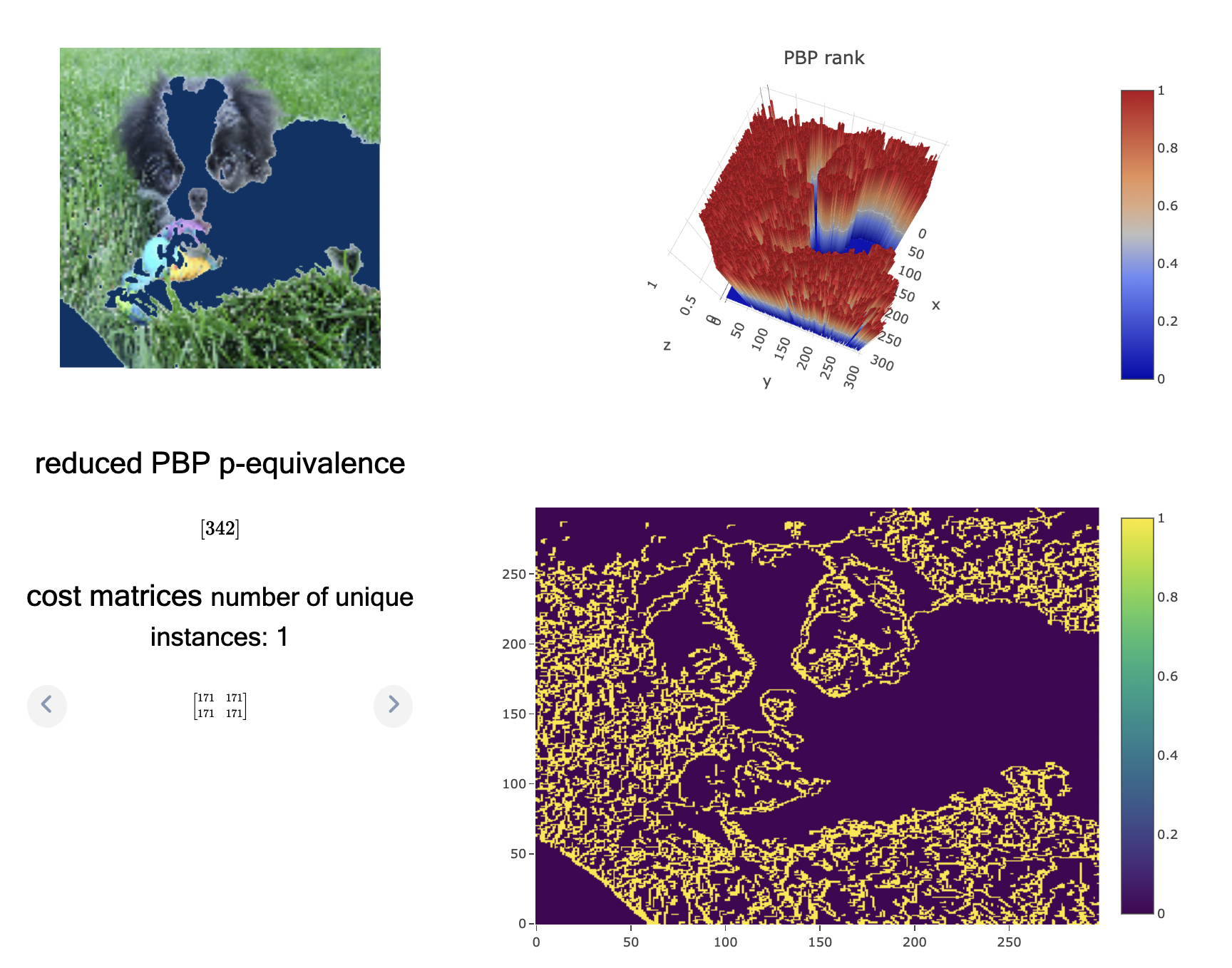}
    \centering
    \caption{Segmentation with all the pre-processing steps included }
    \label{fig:pbp_with_preprocessing}
\end{figure}

We apply our aggregation process on natural images and observe the need for the Gaussian filter as well as the pixel set aggregation preprocessing to achieve useful segmentation. 
Fig.~\ref{fig:input_dog} shows input image of a noisy and natural image that we pass through the segmentation processes without preprocessors as shown in Fig.~\ref{fig:pbp_without_gaussian}, and with processors as shown in Fig.~\ref{fig:pbp_with_preprocessing}.

As can be observed in Fig.~\ref{fig:pbp_with_preprocessing}, a Gaussian preprocessing and pixel grouping step allow us to achieve better edge and blob extraction. 
A Set of pixels, each of size 40, limit our cost range from ${[0, 255]}$ to ${[0, 7]}$ and encourage pronunciation of contrasting regions.

Additionally, we can observe that including a costly operation which aggregates those reduced pseudo-Boolean polynomials into equivalent groups finds numerous, insignificantly small and unuseful groups in the setups which exclude the preprocessing steps, while larger equivalent groups can be aggregated in the setups which involve all the preprocessing steps.

\subsection{Dubai Landscape dataset}

We apply our method with selected processing parameters on the Dubai landscape dataset \cite{loop_semantic_2020}. 
In this experiment we show that our method can extract edges of landscape features on satellite imagery which can be used for semantic segmentation.

Humans in the Loop published an open access dataset annotated for a joint project with the Mohammed Bin Rashid Space Center in Dubai, UAE, which consists of aerial imagery of Dubai obtained by MBRSC satellites and annotated with pixel-wise semantic segmentation in 6 classes \cite{loop_semantic_2020}. 

The full solution on this dataset requires placing labels on each segmented region, however our current method can only segment regions on boundary edges.

The distinctive advantages of our method against the state-of-art neural network based solutions in instance segmentation are:
\begin{itemize}
    \item  blind segmentation(no learning is involved, which can be prone to overfitting/under fitting issues)
    \item faster and CPU friendly segmentation
    \item explainable mathematical steps to segmentation.
\end{itemize}

Our method is also not limited to a given dataset, since it works in a blind manner, which does not require prior familiarity of related image data except for purposes of choosing the threshold ranges. 
Provided with images which contain contrasting features, we can guarantee that our method will propose segmentations of features in pure mathematical and deterministic steps.

\begin{figure}
    \includegraphics[width=0.9\textwidth]{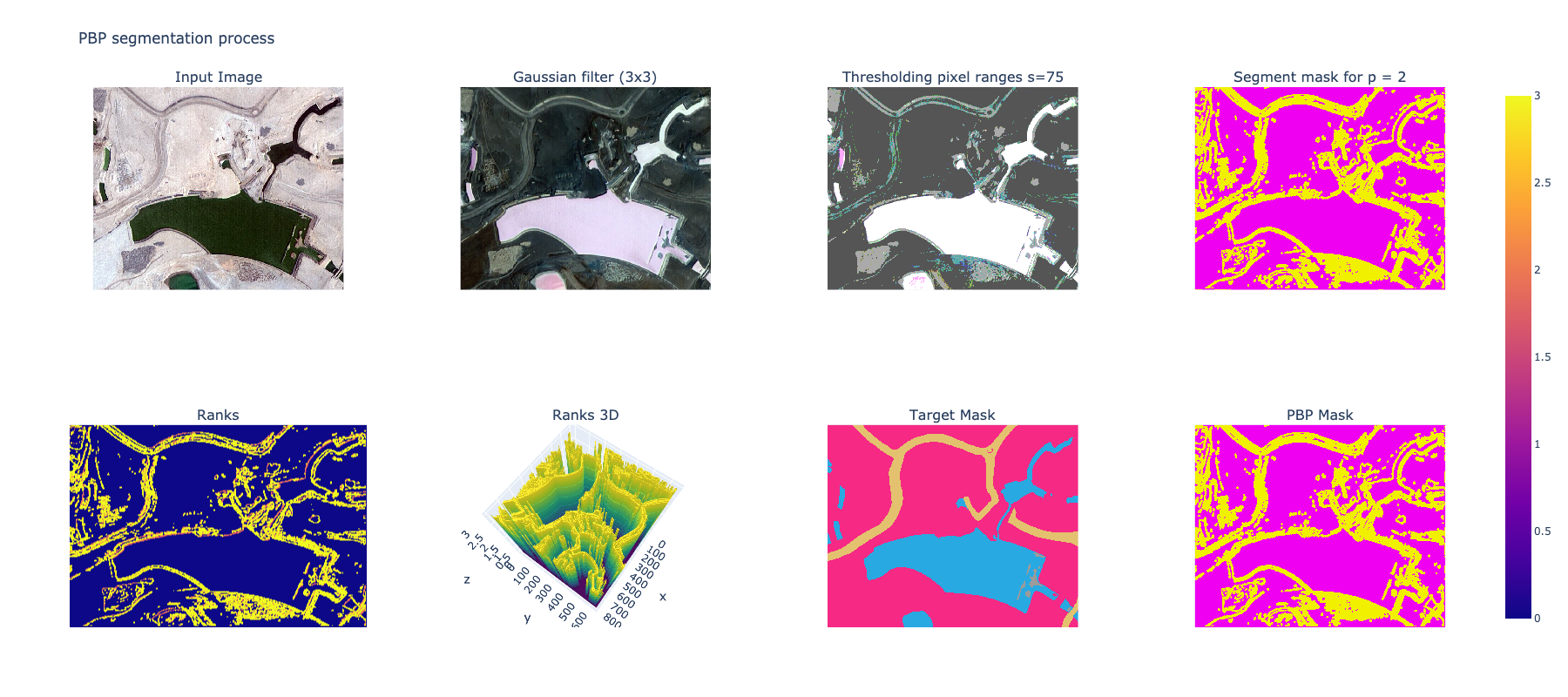}
    \centering
    \caption{Example segmentation [Dubai landscape]}
    \label{fig:dubai_landascape_example}
\end{figure}

Fig.~\ref{fig:dubai_landascape_example} show an example processing of our method on an image sample in the Dubai Landscape dataset \cite{loop_semantic_2020}.

Our method brings to limelight unbiased and fast computer vision in a segmentation task. 
The parameters required to achieve useful segmentation using our method are limited to:
\begin{enumerate}
    \item Gaussian filter kernel size, 
    \item pixel thresholding size and,
    \item patch sizes. 
\end{enumerate} 

The optimal choice of these parameters is the only limitations for the generalization of our proposed method, and we propose in our future work, an automized process for selecting these parameters. 

The performance of the method, based on the choice of the patch size parameter is linearly dependent: the smaller the patch size, the finer the segmentation but longer processing, and vice-versa.

\section{Conclusion}

In this article, we presented our proposed method of formulating pseudo-Boolean polynomials on image patches which results in unsupervised edge detection, blob extraction and image segmentation processes.
We managed to show that our proposed method, works in a fast, unbiased and competitively accurate manner in segmenting contrasting regions in image data. 
We plan to automate the choosing of processing parameters so that our method achieves full functionality as a general purpose image segmentation tool and one of the major tasks in our next challenges is focused on grouping blob regions based on colour histograms to provide labels and consequently achieve complete semantic segmentation.

\section{Acknowledgements}

Tendai Mapungwana Chikake and Boris Goldengorin’s research was supported by Russian Science Foundation project \href{https://rscf.ru/en/project/21-71-30005/}{project No. 21-71-30005}.
 
Boris Goldengorin acknowledges Scientific and Educational Mathematical Center “Sofia Kovalevskaya Northwestern Center for Mathematical Research” for financial support of the present study (agreement No 075-02-2023-937, 16.02.2023)

\bibliographystyle{spmpsci_unsort}
\bibliography{./bib}

\end{document}